\documentclass[sigconf,prologue,table]{acmart}

\usepackage{multirow}
\usepackage{pifont}
\AtBeginDocument{%
  }

\copyrightyear{2025}
\acmYear{2025}
\setcopyright{acmlicensed}\acmConference[Preprint]{ArXiv Version}{2025}{.}
\acmBooktitle{Preprint}
\acmDOI{xxx}
\acmISBN{xxx}

\begin{document}

\title{Sign Spotting Disambiguation using
Large Language Models}

\author{Low Jian He}
\orcid{0009-0009-0452-4374}
\email{jianhe.low@surrey.ac.uk}
\affiliation{%
  \institution{Centre for Vision, Speech and Signal
Processing (CVSSP) \\ University of Surrey}
  \city{Guildford}
  \country{United Kingdom}
}

\author{Ozge Mercanoglu Sincan}
\orcid{0000-0001-9131-0634}
\email{o.mercanoglusincan@surrey.ac.uk}
\affiliation{%
  \institution{Centre for Vision, Speech and Signal
Processing (CVSSP) \\ University of Surrey}
  \city{Guildford}
  \country{United Kingdom}
}

\author{Richard Bowden}
\orcid{0000-0003-3285-8020}
\email{r.bowden@surrey.ac.uk}
\affiliation{%
  \institution{Centre for Vision, Speech and Signal
Processing (CVSSP) \\ University of Surrey}
  \city{Guildford}
  \country{United Kingdom}
}


\begin{abstract}
Sign spotting, the task of identifying and localizing individual signs within continuous sign language video, plays a pivotal role in scaling dataset annotations and addressing the severe data scarcity issue in sign language translation. While automatic sign spotting holds great promise for enabling frame-level supervision at scale, it grapples with challenges such as vocabulary inflexibility and ambiguity inherent in continuous sign streams. Hence, we introduce a novel, training-free framework that integrates Large Language Models (LLMs) to significantly enhance sign spotting quality. Our approach extracts global spatio-temporal and hand shape features, which are then matched against a large-scale sign dictionary using dynamic time warping and cosine similarity. This dictionary-based matching inherently offers superior vocabulary flexibility without requiring model retraining. To mitigate noise and ambiguity from the matching process, an LLM performs context-aware gloss disambiguation via beam search, notably \textit{without fine-tuning}. Extensive experiments on both synthetic and real-world sign language datasets demonstrate our method's superior accuracy and sentence fluency compared to traditional approaches, highlighting the potential of LLMs in advancing sign spotting.
\end{abstract}




\maketitle

\section{Introduction}
\begin{figure}[t]
    \centering
    \includegraphics[width=1\linewidth]{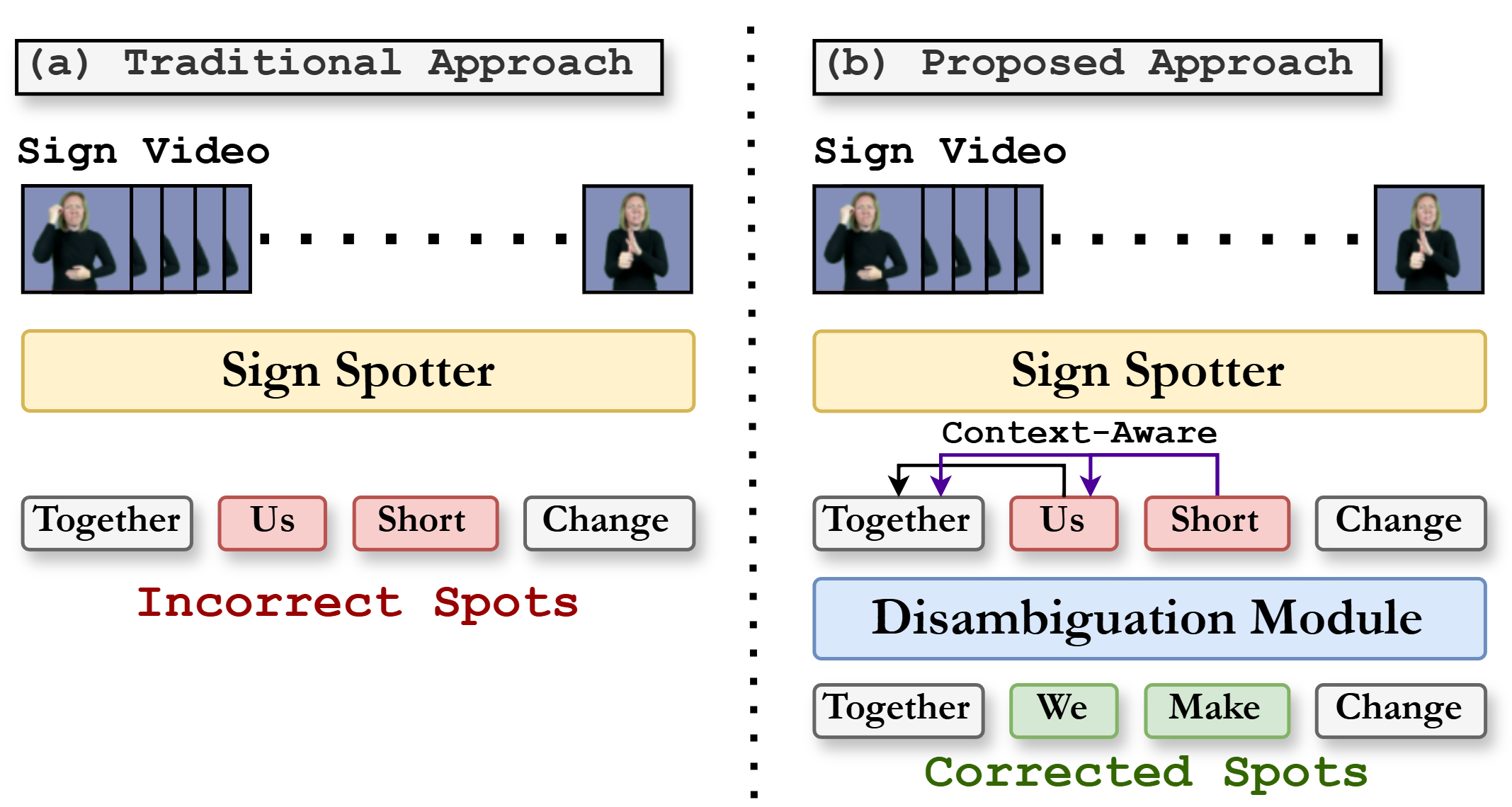}
    \caption{\textbf{System Overview.} (a) Traditional sign spotters classify segments independently. (b) Our proposed approach integrates context-aware disambiguation, leveraging preceding segments for linguistically coherent gloss sequences.}
    \label{fig:teaser}
\end{figure}

Sign languages are complex visual languages expressed through coordinated hand gestures, handshapes, facial expressions, and body posture, governed by distinct grammatical and syntactic rules \cite{signlanguage, whatissignlanguage}. With over 70 million Deaf individuals worldwide using more than 200 distinct sign languages \cite{NUNEZMARCOS2023118993}, developing models to understand sign language at scale is both urgent and impactful. However, the multimodal and multi-channel nature of signing presents significant challenges for modeling, particularly when attempting to relate it to spoken language. Unlike spoken languages, sign languages are not linearly structured; they often rely on simultaneous visual cues, and exhibit fundamentally different linguistic properties, making alignment with spoken counterparts non-trivial.

Among the most widely studied tasks in the field are sign language recognition and translation, due to their direct utility in accessibility and communication \cite{zhou2023gloss, wong2024signgpt, low2025sage}. Prior work has shown that model performance on these tasks improves substantially when trained on gloss-level annotations \cite{chen2022two, zhang2023sltunet, hu2024corrnet+, chen2022simple, he2025hands} or with the aid of large-scale datasets \cite{li2025uni}. Glosses are intermediate spoken language representations aligned at the sign level in continuous sign video; however, their annotations are costly, time-consuming, and requires expertise. For instance, annotating 90 seconds of video in How2Sign took one hour \cite{duarte2021how2sign}. Thus, existing gloss-annotated datasets are small in scale, restricting supervised methods. In contrast, large-scale datasets, pairing videos with spoken language sentences offer broader coverage but lack temporal alignment and structural correspondence; as the order, grammar, and vocabulary of spoken language differ from sign language \cite{Albanie2021BBCOxfordBS, shi2022open, tanzer2024youtube}. This weak supervision introduces ambiguity and hinders translation models from trivially learning precise visual-to-linguistic mappings. 

To bridge this gap, one promising solution is sign spotting, which is the task of localizing and identifying individual signs within continuous signing. While spotting can potentially facilitate scalable supervision, most existing methods are constrained to isolated dictionary look-up schemes \cite{lookingfor, varol2022scale} or hierarchical temporal localization frameworks \cite{Wong2022HierarchicalIF}. Although these methods can identify signs at coarse temporal resolutions, they ignore contextual cues, which are crucial for disambiguating visually similar signs. This is particularly problematic, as subtle differences in facial expressions or hand orientation can lead to drastically different meanings, and context is often the only reliable signal for resolving such ambiguities. 

Motivated by this, we propose a context-aware disambiguation framework that integrates an LLM into the sign spotting pipeline as seen in Figure \ref{fig:teaser}. While our approach builds on dictionary look-up-based spotting, we enhance it by extracting top-$k$ candidate glosses for each localized segment and leveraging the LLM’s next-token prediction capabilities to evaluate gloss sequences based on linguistic coherence. We cast disambiguation as a constrained decoding task, employing beam search to explore high-probability gloss combinations informed by language priors. This enables the system to move beyond isolated sign predictions and instead produce coherent, contextually grounded gloss sequences. To further improve spotting performance, we explore a range of ensemble and fusion strategies for integrating visual cues. Overall, our approach bridges low-level visual recognition with high-level linguistic reasoning and demonstrates that incorporating LLM-based priors can significantly enhance robustness and accuracy in the presence of sign ambiguity. In summary, our main contributions are as follows:
\begin{itemize}
\item We introduce a novel, context-aware and training-free disambiguation module that can be seamlessly integrated into existing sign spotting systems.
\item We evaluate and present several ensemble and fusion techniques to enhance spotting performance.
\item We show that our disambiguation framework substantially reduces word error rate (WER), demonstrating significant gains over baseline spotting architectures.
\end{itemize}

\begin{figure*}[t]
    \centering
    \includegraphics[width=1\linewidth]{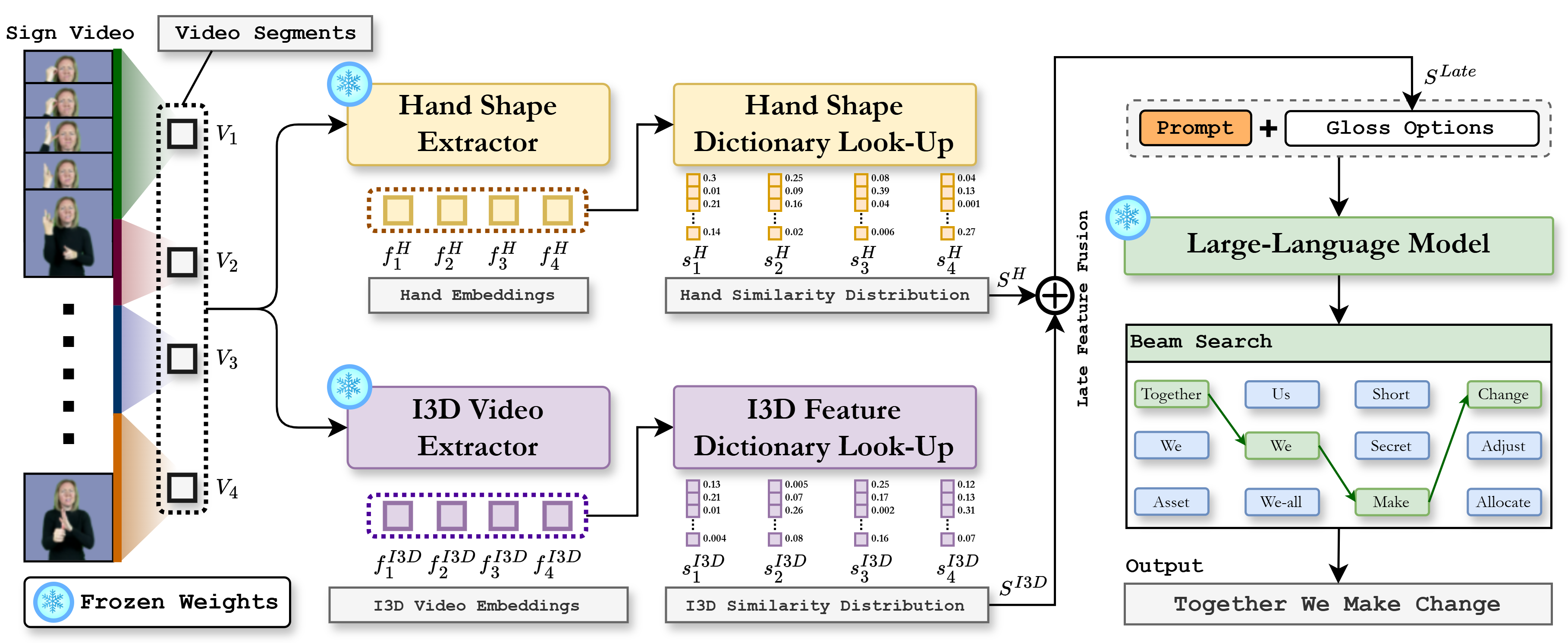}
    \caption{Overall Architecture. Our approach features two main stages: (a) \textbf{Sign Spotting:} Sign segments are first obtained from video. Hand shape and I3D features are then extracted for each segment, and subsequently passed to a dictionary lookup to yield a similarity distribution. Late fusion is then employed to further enhance distinguishability. (b) \textbf{Disambiguation Module:} This module takes the similarity distribution and passes it to an LLM. The LLM's linguistic capabilities are then leveraged, as we extract its transition probabilities and a beam search algorithm uses it  to identify the most coherent sequence of sign glosses.}
    \label{fig:archoverview}
\end{figure*}

\section{Related Works}
\textbf{Sign language spotting} is a key task in sign language understanding, where the goal is to identify the temporal boundaries and the identities of signs within continuous signing videos, given a predefined vocabulary. While underutilized, spotting holds huge potential for scaling the annotation of large sign language datasets. By reducing the cost and effort of manual labeling, it can facilitate the creation of richly supervised data, leading to more accurate and robust sign language translation systems. Beyond annotation, sign spotting also offers further possibilities, including keyword-based video search, dataset curation, and interactive learning applications.

Early approaches to sign spotting largely relied on handcrafted features and traditional sequence modeling techniques. Conditional Random Fields (CRFs), for instance, were employed with adaptive thresholding to differentiate between meaningful signs and co-articulation \cite{yang2008sign}. Other methods modeled the spatial distribution of non-face skin regions using handcrafted histograms, paired with Dynamic Time Warping (DTW) for sequence alignment and retrieval \cite{viitaniemi2014s, Berndt1994UsingDT}. In parallel, spatio-temporal patterns such as Sequential Interval Patterns (SIPs) were explored as sign representations, with work proposing a hierarchical forest of SIP trees to enhance spotting robustness across varying sign instances \cite{ong2014sign}.

The deep learning era then introduced more scalable and accurate methods. One-shot spotting approaches emerged by computing similarity between embeddings of isolated signs and continuous video segments, leveraging powerful video encoders like the Inflated 3D ConvNet (I3D) \cite{carreira2017quo}. For example, \cite{Momeni2020WatchRA} proposed a multi-supervision framework using an I3D, and spotted signs via computing similarity scores between dictionary embeddings and localized segments. Meanwhile, the Sign-Lookup approach \cite{lookingfor} utilized a 3D Convolutional Neural Network (CNN) with a Transformer, performing a sliding window over the video input to compute similarities with dictionary signs. More recently, dictionary-free approaches have also emerged, with \cite{Wong2022HierarchicalIF} introducing a hierarchical framework that uses I3D features at different layers to learn coarse-to-fine temporal boundaries and perform sign classifications.


\section{Methodology}
As illustrated in Figure \ref{fig:archoverview}, our proposed sign spotting disambiguation framework introduces a novel integration of LLMs into the sign spotting pipeline to enhance the selection of sign candidates through linguistic reasoning. The idea is to treat the probability distribution output from a sign spotting model as a set of candidate glosses, which are then passed to an LLM that acts as a contextual scorer. By leveraging the LLM’s strong language modeling and next-token prediction capabilities, we reinterpret its output as transition probabilities within a beam search decoding process.

While the LLM introduces linguistic priors into the pipeline, the quality of the spotting predictions (output probabilities) remains critical. Thus, we investigate various strategies for improving these predictions in Section \ref{feature fusion}, including ensembling multiple sign spotting outputs and applying feature-level fusion to enhance the robustness and precision of the extracted gloss candidates.


\subsection{Sign Spotter}

Our sign spotter consists of three main components: (1) a multi-branch \textbf{feature extraction module}, where each stream is responsible for extracting distinct aspects of the signing input; (2) a \textbf{dictionary-based matching module}, which performs candidate retrieval based on embedding similarity; and (3) a \textbf{feature fusion stage}, where we explore various ensembling strategies to integrate motion and hand-specific information.

\subsubsection{\textbf{Feature Extraction}}

We employ two specialized neural networks to extract complementary representations from the signing video. The first is a spatiotemporal encoder based on the \textit{\textbf{Inflated 3D ConvNet (I3D)}} \cite{carreira2017quo}, which captures coarse motion patterns and global appearance cues. We employ the I3D which was pretrained on the \textit{BOBSL} dataset \cite{Albanie2021BBCOxfordBS}, and finetuned for sign language recognition by \cite{sincan2024using}. Here, each video is processed using a sliding window of fixed length to extract local segment-level embeddings. 

Formally, we define the input video as, $V \in \mathbb{R}^{T \times H \times W \times 3}$, where $T$ is the total number of frames, and $H \times W$ is the spatial resolution. We extract features using overlapping 16-frame windows with stride 1. For each window $V_{t:t+15}$, the I3D network produces a 1024-dimensional embedding:

\[
\mathbf{f}^{\text{I3D}}_t = \text{I3D}(V_{t:t+15}) \in \mathbb{R}^{1024}
\]

Aggregating over the entire sequence, we obtain the motion-based feature matrix:

\[
\mathbf{F}^{\text{I3D}} = \left[ \mathbf{f}^{\text{I3D}}_1, \mathbf{f}^{\text{I3D}}_2, \ldots, \mathbf{f}^{\text{I3D}}_{T-15} \right]^\top \in \mathbb{R}^{(T - 15) \times 1024}
\]

As the I3D uses 3D convolutions, each frame will always be influenced by its temporal context, which allows the model to capture rich motion and appearance patterns that evolve over time. This resulting feature sequence $\mathbf{F}^{\text{I3D}}$ thus serves as a temporally-aware embedding of the input video, which is later fused with complementary features for candidate retrieval.

In parallel, we employ a \textit{\textbf{ResNeXt-101-based hand shape encoder}} to capture fine-grained spatial cues from both hands. This model is similar to DeepHand \cite{7780781}, which was originally trained on the 1 Million Hand Images dataset for handshape classification. In our adaptation, we leverage a deeper ResNeXt-101 backbone, pretrained on the same dataset for 60-way handshape classification. However, we remove the classification head and instead use the penultimate layer features as dense hand embeddings.

To isolate hand regions from full-frame signing videos, we first detect upper-body pose landmarks using \textit{MediaPipe} \cite{DBLP:journals/corr/abs-1906-08172}, and then crop bounding boxes around the left hands (LH) and right hands (RH) at each frame. Thus, given an input video $V \in \mathbb{R}^{T \times H \times W \times 3}$, MediaPipe produces localized crops:

\[
H^{\text{LH}}_t, H^{\text{RH}}_t \in \mathbb{R}^{h \times w \times 3}, \quad \forall t \in \{1, \dots, T\}
\]

where $H^{\text{LH}}_t$ and $H^{\text{RH}}_t$ denote the cropped left and right hand regions at time step $t$. These are then passed through the ResNeXt-101 encoder to extract hand-specific features:

\[
\mathbf{F}^{\text{LH}}_t = \text{ResNeXt}(H^{\text{LH}}_t), \quad \mathbf{F}^{\text{RH}}_t = \text{ResNeXt}(H^{\text{RH}}_t) \in \mathbb{R}^{2048}
\]

These embeddings capture high-resolution spatial information from each hand independently, providing a complementary signal to the more holistic, motion-oriented features extracted by the I3D.

\subsubsection{\textbf{Dictionary-Based Matching}}

To obtain gloss candidates, we employ a dictionary-based matching approach that leverages feature similarity. While alternative methods, such as those proposed by \cite{Wong2022HierarchicalIF, sincan2024using}, forgo dictionaries by using pretrained classifiers (e.g., a final fully connected layer for prediction), they are inherently constrained by a fixed vocabulary size. This rigidity, though potentially offering robustness for predefined signs, leads to significant Out-Of-Vocabulary (OOV) issues when applied to novel data or signs not encountered during training. Such inflexibility is a critical drawback, as a primary application of sign spotting is the annotation of large-scale datasets with potentially dynamic or undefined vocabularies. In contrast, dictionary-based methods offer the key advantage of allowing new vocabulary items to be incorporated into the lookup table without necessitating model retraining, thereby effectively addressing OOV instances. Thus, we adopt the dictionary-based approach due to its flexibility and further explore its performance benefits in the results of Sec \ref{sec:full sys eval}.

Our sign dictionary is constructed from isolated British Sign Language (BSL) samples and comprises of 1,000 vocabulary items (details at Sec. \ref{sec:full sys eval}). Each dictionary entry encodes the prototypical visual representation of a gloss using extracted feature embeddings. Formally, we define the dictionary as:
\begin{equation}
    \mathcal{D} = \left\{ (\mathbf{D}_i, g_i) \right\}_{i=1}^{1000},
\end{equation}
where $\mathbf{D}_i$ is the feature embedding for the $i$-th gloss and $g_i$ is its corresponding label.

Each dictionary embedding $\mathbf{D}_i$ is composed by concatenating the I3D motion features and the hand shape features extracted from both left and right hands:
\begin{equation}
    \mathbf{D}_i = \mathbf{F}^{I3D}_{i} \oplus \mathbf{F}^{LH}_{i} \oplus \mathbf{F}^{RH}_{i} \in \mathbb{R}^{5120},
\end{equation}
where $\oplus$ denotes feature concatenation.

This dictionary is constructed using isolated sign videos to ensure clean visual representations. Importantly, the structure of $\mathbf{D}_i$ is modular, allowing us to build variant-specific dictionaries tailored to different feature configurations \textit{without retraining}. For instance, when exploring fusion approaches (Sec. \ref{feature fusion}), we constructed separate dictionaries using only I3D features and right-hand embeddings due to the difference in feature extraction pipeline.

\paragraph{\textbf{Similarity Computation.}}
For each sign unit $U_x$, where $x \in \{1, \dots, X\}$ and $X$ is the total number of candidate segments from a continuous signing video, we compute similarity scores against all dictionary entries $\mathbf{D}_i$ using two complementary metrics: (i) \textbf{Dynamic Time Warping (DTW)}~\cite{muller2007dynamic}, which aligns frame-level features of $U_x$ and $\mathbf{D}_i$ to capture temporal structure while accounting for variations in signing speed; and (ii) \textbf{Cosine Similarity}, which is computed between pooled segment-level embeddings. Specifically, frame-wise features are temporally pooled into fixed-size vectors $\mathbf{F}_x \in \mathbb{R}^d$, enabling efficient segment-level comparison.



The final similarity score is then computed using a weighted sum as seen in equation \ref{final similarity}, and visualized in Figure \ref{fig:latefuse}:
\begin{equation}
\label{final similarity}
    \text{score}(U_x, \mathbf{D}_i) = (\alpha_{\text{s}} - 1) \cdot \text{sim}_{\text{DTW}}(U_x, \mathbf{D}_i) + \alpha_{\text{s}} \cdot \text{sim}_{\text{cos}}(U_x, \mathbf{D}_i),
\end{equation}
where $\alpha_{\text{s}} \in [0,1]$ is a hyperparameter controlling the contribution of each similarity metric. Note that DTW is a distance metric (lower means more similarity); thus, we multiply its scalar with a negative sign to ensure higher values consistently indicate greater similarity.

\subsubsection{\textbf{Feature Fusion}} \label{feature fusion}

To enhance the quality of gloss selection within the Top-k candidate set, we investigate several feature fusion strategies aimed at increasing the discriminative power of the segment representations. Specifically, we explore: (i) \textbf{late fusion}, where features are combined after being independently encoded; (ii) \textbf{intermediate fusion}, which integrates features within a shared embedding space during encoding; and (iii) a \textbf{full-ensemble} strategy that jointly leverages both approaches. These fusion mechanisms are crucial for ensuring that the LLM receives more informative gloss candidates for downstream disambiguation.

\label{featurefusion}

\begin{figure}[t]
    \centering
    \includegraphics[width=1\linewidth]{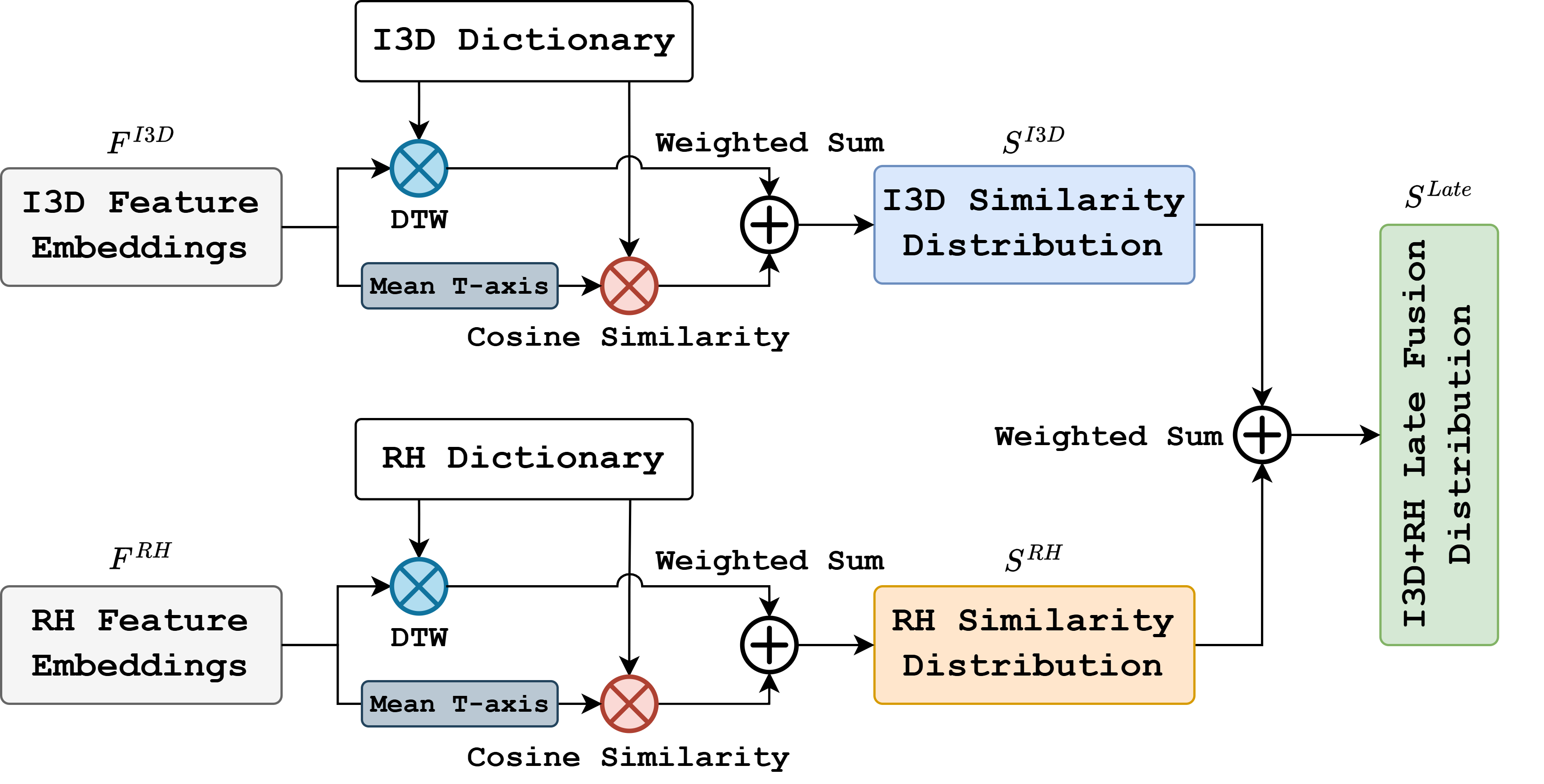}
    \caption{Illustration of Dictionary Look-Up and Late Fusion.}
    \label{fig:latefuse}
\end{figure}

\paragraph{\textbf{Late Fusion}}
For late fusion, we integrate the independently computed similarity distributions derived from the I3D and RH embeddings. Initially, as depicted in Figure \ref{fig:latefuse}, the I3D embedding, $\mathbf{F}^{\text{I3D}}$, is evaluated against its dedicated I3D dictionary. This evaluation leverages the similarity computation defined in equation \ref{final similarity}, resulting in a similarity distribution $\mathbf{S}^{\text{I3D}} \in \mathbb{R}^{V}$, where $V$ represents the total number of vocabulary entries. In parallel, the RH embedding, $\mathbf{F}^{\text{RH}}$, undergoes an analogous process with its corresponding RH dictionary to generate $\mathbf{S}^{\text{RH}}$.

The final late fusion of these distributions is then achieved through a weighted summation:
\begin{equation} \label{eq:late_fusion}
\mathbf{S}^{\text{Late}} = \alpha_{\text{late}} \cdot \mathbf{S}^{\text{I3D}} + (1 - \alpha_{\text{late}}) \cdot \mathbf{S}^{\text{RH}},
\end{equation}
where $\alpha_{\text{late}} \in [0,1]$ is a tunable hyperparameter that balances the influence of the I3D and RH shape features. We observed through preliminary experimentation that the inclusion of LH features provided negligible performance improvements; thus, they are excluded from this specific late fusion strategy.

\begin{figure}[b]
    \centering
    \includegraphics[width=1\linewidth]{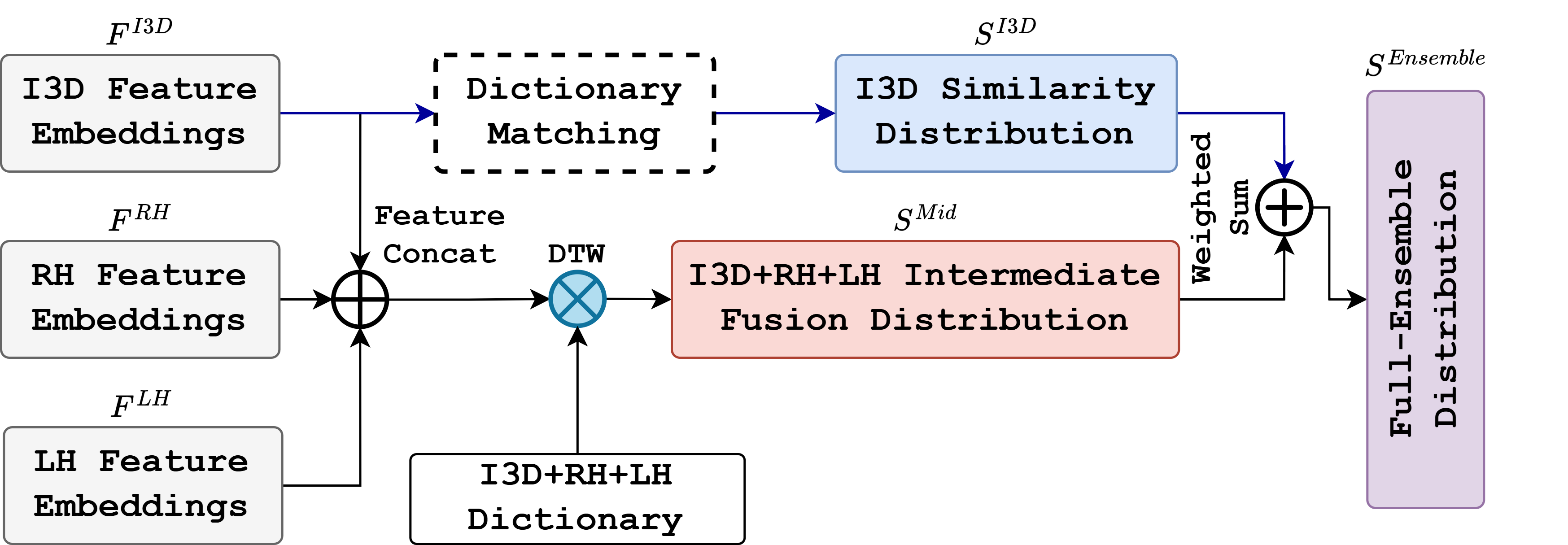}
    \caption{Illustration of Mid Fusion and Full-Ensemble.}
    \label{fig:interfuse}
\end{figure}

\paragraph{\textbf{Intermediate Fusion}} 
For intermediate fusion, we combine feature embeddings from multiple encoders before similarity computation. Specifically, for each sign unit, the I3D ($\mathbf{F}^{\text{I3D}} \in \mathbb{R}^{1024}$), RH ($\mathbf{F}^{\text{RH}} \in \mathbb{R}^{2048}$), and LH ($\mathbf{F}^{\text{LH}} \in \mathbb{R}^{2048}$) embeddings are concatenated to form a unified feature vector $\mathbf{F}^{\text{Mid}}$:

$$\mathbf{F}^{\text{Mid}} = \mathbf{F}^{\text{I3D}} \oplus \mathbf{F}^{\text{RH}} \oplus \mathbf{F}^{\text{LH}} \in \mathbb{R}^{5120}$$

This fused vector then serves as the input for dictionary-based matching, as depicted in Figure \ref{fig:interfuse}. This process ultimately yields the intermediate fusion similarity distribution $\mathbf{S}^{\text{Mid}} \in \mathbb{R}^{V}$. Note that unlike late fusion, we exclusively employ DTW for similarity computation at this stage; as cosine similarity, which typically involves temporal averaging, was found to dilute crucial modality-specific information post-concatenation, leading to suboptimal performance.

\paragraph{\textbf{Full-Ensemble}}
For the final full-ensemble strategy, we combine the concepts of both the late and intermediate fusion approaches. Specifically, we perform a weighted late fusion between the similarity distribution obtained from intermediate fusion, $\mathbf{S}^{\text{Mid}}$, and the I3D-specific distribution, $\mathbf{S}^{\text{I3D}}$, as depicted in Figure \ref{fig:interfuse}. The ensemble similarity distribution, $\mathbf{S}^{\text{Ensemble}}$, is computed as:
\begin{equation} \label{eq:ensemble}
\mathbf{S}^{\text{Ensemble}} = \alpha_{\text{ens}} \cdot \mathbf{S}^{\text{Mid}} + (1 - \alpha_{\text{ens}}) \cdot \mathbf{S}^{\text{I3D}},
\end{equation}
where $\alpha_{\text{ens}} \in [0,1]$ is a tunable hyperparameter balancing the contributions of the intermediate and I3D feature modalities.

\subsection{Linguistic Disambiguation} \label{disambiguation section}

This section details the LLM-based disambiguation stage, which comprises of three key components: (i) \textbf{Gloss Candidate Generation}, detailing how initial glosses are prepared from similarity distributions; (ii) \textbf{Prompt Formulation}, describing the structure and content of the prompt used for the LLM; and (iii) \textbf{Beam Search Decoding}, which outlines the final disambiguation process.

\subsubsection{\textbf{Gloss Candidate Generation}}
\begin{figure}[t]
    \centering
    \includegraphics[width=1\linewidth]{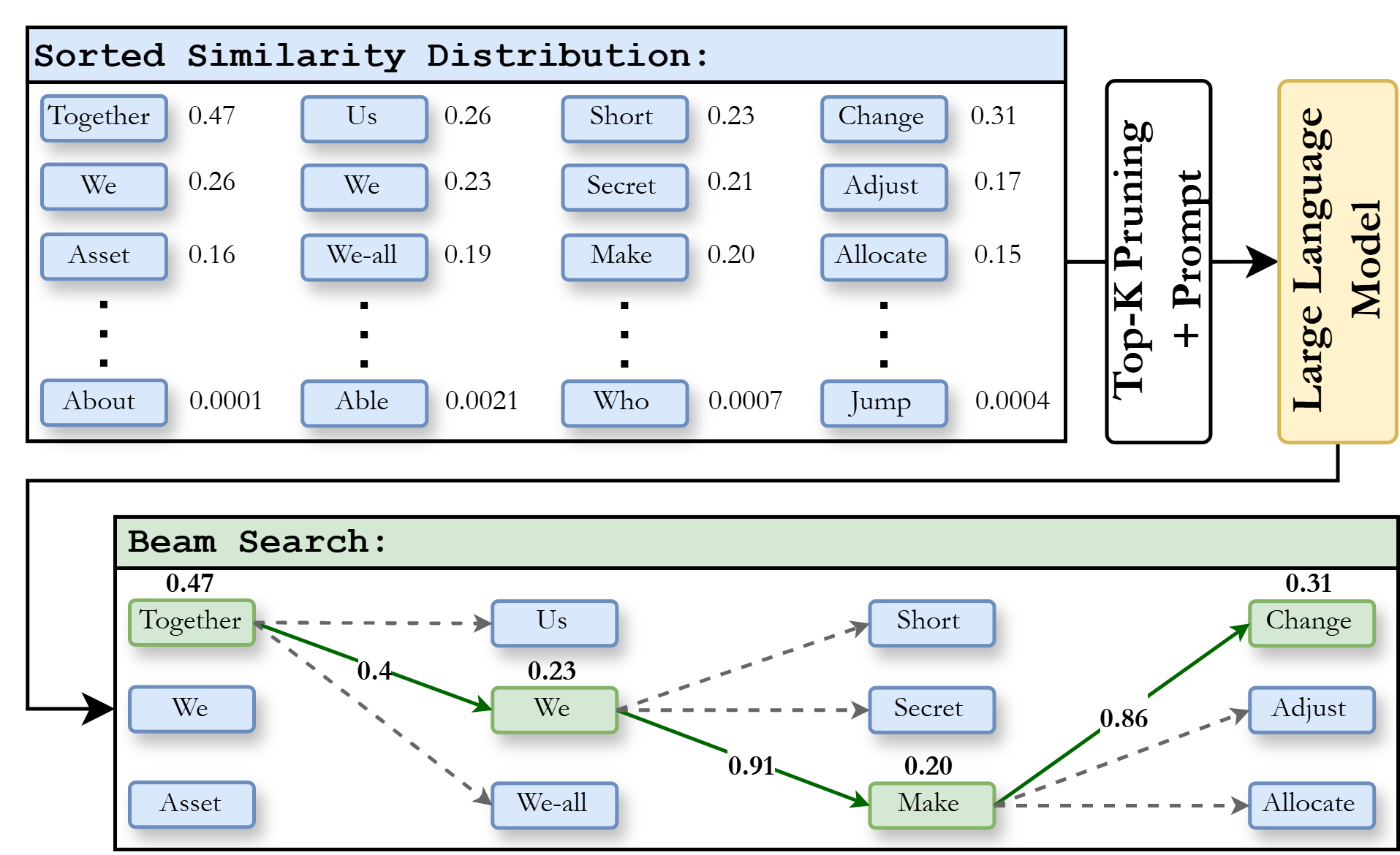}
    \caption{Visualization of the Disambiguation Process.}
    \label{fig:beamsearch}
\end{figure}

From the preceding dictionary look-up stage, a \textit{similarity distribution} $\mathbf{S} \in \mathbb{R}^{V}$ (where $V$ is the vocabulary size) is obtained for each sign segment, indicating the relevance of each potential gloss candidate. This raw distribution undergoes normalization via a softmax function, and its Top-$k$ gloss candidates are then selected and forwarded to the LLM along with the prompt for subsequent disambiguation (Figure \ref{fig:beamsearch}).

\begin{figure}[H]
    \centering
    \includegraphics[width=1\linewidth]{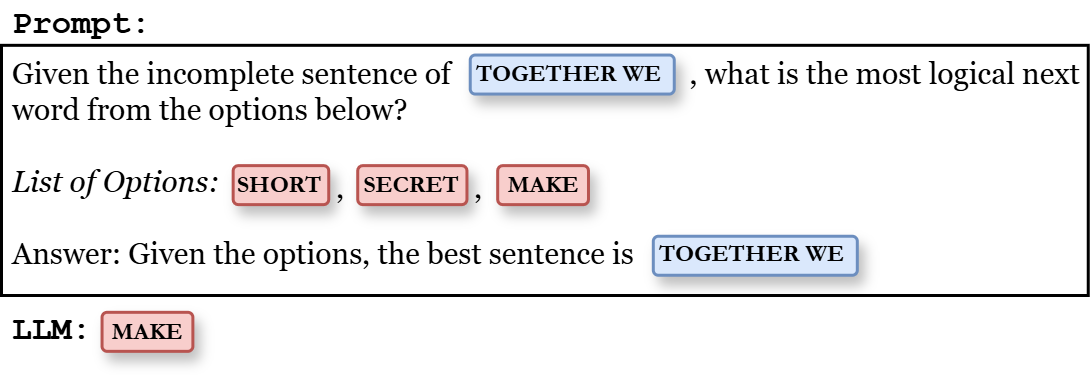}
    \caption{Example of Our Prompt Template.}
    \label{fig:prompt}
\end{figure}

\subsubsection{\textbf{Prompt Formulation}}
The prompt provided to the LLM is dynamically constructed to incorporate both contextual information and candidate glosses. As seen in Figure \ref{fig:prompt}, specific elements of the prompt are instantiated with variable content. For instance, the red-highlighted "Short," "Secret," and "Make" represent the Top-$k$ gloss candidates, while the blue-highlighted "Together we" signifies the previous sign glosses, providing contextual information.

Thus, the prompt's core function is to leverage the LLM's predictive capabilities for sentence completion using the provided context and gloss options (e.g., completing "Together we \_\_\_"). To extract the transition probabilities for each candidate, the prompt is specifically designed to compel the LLM to predict one of the gloss options as its next word; thus, allowing for direct extraction of each candidate's probability as the subsequent token in the sequence.

\begin{table*}[tbp]
\caption{Hyperparameter Optimization of the $\alpha$ from ranges 0 to 1 for Different Fusion Equations based on Top-K Accuracy}
\centering
\label{table:tune1}
\begin{tabular}{@{}ccccccccccccccccc@{}}
\toprule
\multicolumn{1}{c}{Equation} & ${\text{Accuracy}\uparrow}$ & \multicolumn{1}{c}{$\alpha=0$} & \multicolumn{1}{c}{$\alpha=0.1$} & \multicolumn{1}{c}{$\alpha=0.2$} & \multicolumn{1}{c}{$\alpha=0.3$} & \multicolumn{1}{c}{$\alpha=0.4$}    & \multicolumn{1}{c}{$\alpha=0.5$} &  \multicolumn{1}{c}{$\alpha=0.6$} & \multicolumn{1}{c}{$\alpha=0.7$} & \multicolumn{1}{c}{$\alpha=0.8$} & \multicolumn{1}{c}{$\alpha=0.9$} & \multicolumn{1}{c}{$\alpha=1.0$}  \\ \midrule
\multirow{2}{*}{$\mathbf{S}^{\text{I3D}}$, Eq. \ref{final similarity}} & Top-5 & 0.7888 & 0.8141 & 0.8155 & \textbf{0.8272} & 0.8255 & 0.8222  & 0.8222 & 0.8166 & 0.8066 & 0.8103 & 0.8056 \\
& Top-1 & 0.4605 & 0.4788 & 0.4834 & \textbf{0.5151} & 0.5136 & 0.5050  & 0.5050 & 0.5050 & 0.5050 & 0.5136 & 0.5136 \\
\multirow{2}{*}{$\mathbf{S}^{\text{RH}}$, Eq. \ref{final similarity}}  & Top-5 & 0.5511 & 0.5608 & 0.5670 & 0.5725 & 0.5917 & \textbf{0.5965} & \textbf{0.5965} & 0.5923 & 0.5923 & 0.5875 & 0.5917 \\
& Top-1 & 0.3212 & 0.3242 & 0.3200 & 0.3219 & 0.3219 & 0.3261  & 0.3280 & 0.3280 & 0.3338 & \textbf{0.3397} & \textbf{0.3397} \\
\multirow{2}{*}{$\mathbf{S}^{\text{Late}}$, Eq. \ref{eq:late_fusion}}  & Top-5 & 0.5923 & 0.6558 & 0.6973 & 0.7625 & 0.7738 & 0.8054 & 0.8132 & 0.8215 & 0.8259 & \textbf{0.8276} & 0.8272 \\
& Top-1 & 0.3338 & 0.4186 & 0.4793 & 0.5122 & 0.4952 & \textbf{0.5451}  & 0.5343 & 0.5259 & 0.5148 & 0.5020 & 0.5151  \\

\multirow{2}{*}{$\mathbf{S}^{\text{Ensemble}}$, Eq. \ref{eq:ensemble}} & Top-5 & 0.8272 & 0.8230 & 0.8197 & 0.8317 & 0.8373 & \textbf{0.8415} & \textbf{0.8415} & \textbf{0.8415} & 0.8393 & 0.8223 & 0.8260 \\
& Top-1 & 0.5151 & 0.5039 & 0.5095 & 0.5137 & 0.5176 & 0.5109 & \textbf{0.5190} & 0.5160 & 0.5080 & 0.5118 & 0.5023\\ \bottomrule 
\end{tabular}
\end{table*}

\subsubsection{\textbf{Beam Search Decoding}}
For the final disambiguation, a beam search algorithm is employed to combine the two critical probability types: \textbf{\textit{emission probabilities}} (representing gloss candidates from the sign spotter) and \textbf{\textit{transition probabilities}} (reflecting linguistic coherence from the LLM).

Thus, for the $x^{\text{th}}$ sign unit, where $x \in \{1, \dots, X\}$ ($X$ denoting the total number of units), we first obtain a set of top-$k$ gloss candidates $\mathcal{C}_x = \{(g_x^j, s_x^j)\}_{j=1}^k$. Here, $s_x^j$, which is derived from the pre-computed dictionary similarity (e.g., $\mathbf{S}^{\text{Late}}$, $\mathbf{S}^{\text{Mid}}$, or $\mathbf{S}^{\text{Ensemble}}$), serves as the emission probability for candidate gloss $g_x^j$. Meanwhile, the LLM provides our transition probabilities $p(g_x | g_{1:x-1})$, indicating the likelihood of gloss $g_x$ given the sequence of previously selected glosses $g_{1:x-1}$, providing us with linguistic guidance.

Beam search then aims to find the optimal gloss sequence $\hat{g}_{1:X}$ by maximizing a combined score that leverages these probabilities:

\begin{equation} \label{eq:beam_search_objective}
\hat{g}_{1:X} = \underset{g_{1:X} \in \prod_{x=1}^{X} \mathcal{C}_x}{\arg\max} \sum_{x=1}^{X} \left( \log p(g_x | g_{1:x-1}) + \alpha_{\text{bs}} s_x \right)
\end{equation}

In this objective, $\prod_{x=1}^{X} \mathcal{C}_x$ denotes the Cartesian product of the candidate sets. The conditional probability $p(g_x | g_{1:x-1})$ is derived from the frozen LLM's logits, while the term $s_x$ represents the emission probability for the selected gloss $g_x$ at the $x^{\text{th}}$ sign unit. The hyperparameter $\alpha_{\text{bs}}$ weighs the the LLM's linguistic score and the emission probability. To sufficiently explore the hypothesis space, we employ the algorithm with a beam width (BW) of $5$.



\section{Experiments and Results}

Our evaluation of the proposed system employs a two-tiered strategy. First, an \textbf{\textit{isolated gloss disambiguation}} evaluation validates the processes detailed in Sec. \ref{disambiguation section} on synthetic data to ensure the module's efficacy prior to integration. Subsequently, for the \textbf{\textit{full system evaluation}}, the validated disambiguation module is integrated with a sign spotter, and assessed on continuous sign videos and a large sign dictionary comprising 1000 vocabulary entries.

\subsection{Isolated Gloss Disambiguation Evaluation}
\paragraph{\textbf{Synthetic Data Generation for Disambiguation Evaluation}}
To evaluate the gloss disambiguation process in isolation and conduct comprehensive ablation studies, we utilize synthetically generated data. This pipeline is designed to simulate the complexities and ambiguities inherent in real-world sign spotting, encompassing dictionary construction, pseudo-gloss sentence generation, and controlled noise injection to mimic realistic emission probabilities.

\subsubsection{\textbf{Synthetic Dictionary Construction}} \label{dictionary gen}
Existing large-scale British Sign Language (BSL) dictionaries often lack crucial, high-frequency vocabulary. To address this, we construct a synthetic BSL-based dictionary informed by word frequency research. We leverage a word frequency list derived from the Google Web Trillion Word Corpus \cite{engwordfreq}, cross-referencing it with findings from a BSL lexical frequency study \cite{FENLON2014187}, which identified the 100 most frequent BSL signs, and BSLDict \cite{Momeni2020WatchRA}. We thus constructed four vocabulary lists of varying sizes: 1500, 2000, 3000, and 4373 words. The maximum size was capped at 4373, as it corresponded to the highest number of overlapping words between the Google frequency list and the BSLDict sign vocabulary. This constructed dictionary also serves as a template for potential future sign spotting applications.

\subsubsection{\textbf{Pseudo-Gloss Generation}}
To generate synthetic sign spotted units, we employ a pseudo-gloss generation strategy. We first curate a corpus of random English sentences using a random sentence generator \cite{randomword}. Motivated by recent work in sign language translation \cite{wong2024signgpt}, we then create pseudo-glosses by applying Part-of-Speech (POS) tagging via the SpaCy library. Only tokens tagged as ``NOUN'', ``NUM'', ``ADV'', ``PRON'', ``PROPN'', ``ADJ'', and ``VERB'' are retained, simulating the typical lexical content of sign glosses.

\subsubsection{\textbf{Simulating Ambiguity via Noise Injection}}
To realistically simulate the dictionary-matching similarity distributions obtained from visual sign spotters, we initially embed both the synthetic sentences and dictionary vocabulary items using the FastText library and perform direct matching. However, we recognize that this naive approach yields significantly better matches than complex visual embedding; thus, we incorporate two types of noise augmentation to introduce real-life ambiguity:

\paragraph{\textbf{Word Replacement (WR):}} The top-$1$ option of the similarity distribution is replaced with another random word at a specified probability, simulating an incorrect classification by a sign spotter.

\paragraph{\textbf{Distribution Corruption (DC):}} From a given distribution, we randomly select $k$ semantically dissimilar words. We then artificially increase their similarity scores to a value exceeding the highest pre-existing score. This introduces significant noise, ensuring that random, irrelevant words appear in the top-$k$ candidates, thereby stringently testing the disambiguation capability of our method.

\subsubsection{\textbf{Evaluation and Results}}

Based on the methods above, we conduct a series of \textbf{hyperparameter optimization} (HPO) and \textbf{ablation studies} to select the optimal configuration for the disambiguation module and evaluate its overall effectiveness.

\paragraph{\textbf{Hyperparameter Optimization}}
We first perform HPO on the weighted sum parameters used in both the similarity computation and feature fusion stages. Specifically, we tune the balancing hyperparameters $\alpha \in [0,1]$ for four distinct combination strategies: (i) the sum of DTW and Cosine Similarity for \textbf{I3D features} (Eq. \ref{final similarity}, used for $\mathbf{S}^{\text{I3D}}$ derivation), (ii) the sum of DTW and Cosine Similarity for \textbf{RH features} (Eq. \ref{final similarity}, used for $\mathbf{S}^{\text{RH}}$ derivation), (iii) the \textbf{late fusion} strategy (Eq. \ref{eq:late_fusion}), and (iv) the \textbf{full-ensemble} method (Eq. \ref{eq:ensemble}). Performance is measured using Top-1 and Top-5 accuracy.

Based on table \ref{table:tune1}, which summarizes the HPO results, we find that the I3D similarity computation benefits from a greater contribution from DTW. In contrast, the RH similarity computation achieves its peak Top-1 accuracy at $\alpha_{\text{s}}=0.9$, indicating stronger reliance on cosine similarity. For feature fusion, the late fusion strategy shows optimal Top-5 accuracy (0.8276) when I3D features contribute more ($\alpha_{\text{late}}=0.9$). Meanwhile, the full-ensemble method also performs strongly, with peak performances observed at $\alpha_{\text{ens}}=0.6$, suggesting a slightly greater emphasis on the intermediate fusion distribution ($\mathbf{S}^{\text{Mid}}$). Overall, the explicit feature fusion methods consistently outperform the vanilla $\mathbf{S}^{\text{I3D}}$ and $\mathbf{S}^{\text{RH}}$ baselines.

\paragraph{\textbf{Dictionary Size}}
As detailed in Section \ref{dictionary gen}, our synthetic dictionary generation yielded varying vocabulary sizes. Here, we analyze performance across dictionary sizes of 1500, 2000, and 4373 words under different noise augmentation conditions (Table \ref{table:DictResults}). These conditions include Word Replacement (WR) rates of 50\% and 100\%, and Distribution Corruption (DC) rates of 5, 10, and 15. Consistently, the dictionary size of 4373 achieves the lowest Word Error Rates (WER) across all augmentation conditions, demonstrating that a larger vocabulary enhances the sign spotting and disambiguation process due to a broader range of potential matches.

\begin{table}[h]
\caption{Comparisons between Different Dictionary Sizes}
\centering
\label{table:DictResults}
\begin{tabular}{@{}cccccc@{}}
\toprule
WR & DC & D1500$_{\text{WER}\downarrow}$ & D2000$_{\text{WER}\downarrow}$ &  \textbf{D4373$_{\text{WER}\downarrow}$} \\ \midrule
50\%  & 5  & 0.4812   & 0.4749 & \textbf{0.3824}  \\
100\%  & 5  & 0.7670   & 0.7691 & \textbf{0.6206}   \\
50\%  & 10  & 0.4606   & 0.4432 & \textbf{0.3723}   \\
100\% & 10  & 0.8110   & 0.7817 & \textbf{0.6557}  \\ 
50\%  & 15  & 0.5043   & 0.4585 & \textbf{0.4041}   \\
100\% & 15  & 0.8111   & 0.7891 & \textbf{0.7163}  \\
\bottomrule
\end{tabular}
\end{table}

\paragraph{\textbf{Augmentation Strength}} Additionally, we also evaluate the robustness of our approach to noise and conduct an ablation study comparing different LLMs. Specifically, we compare \textit{Phi-3 Mini} \cite{phi3}, selected for its strong performance among lightweight models, with \textit{Gemma-2 9B} \cite{gemma_2024}, the most capable model that fit our RTX3090 GPU. Our evaluation includes tests with increasing Word Replacement (WR) rates (25\% to 100\%) and Distribution Corruption (DC) levels (5 to 20), detailed in Table \ref{table:WRresults}, alongside extreme conditions (DC=30; WR=50\%, 100\%) with varying Beam Widths, presented in Table \ref{table:WRLCresult}.

In Table \ref{table:WRresults}, we see that Gemma-2 outperformed Phi-3 across all scenarios, highlighting the benefit of utilizing a larger LLM. However, overall, both models demonstrated clear disambiguation capabilities, even at 100\% WR, as they successfully reduced Top-1 WER to under 0.63 ($100\% \rightarrow < 63\%$). This suggests that successful corrections are achievable as long as target glosses remain within the top candidates. However, Phi-3 struggled significantly at high DC (DC=20), yielding a WER of 0.44. This performance, offering only a 6\% reduction in WER ($50\% \rightarrow 44\%$), highlights its sensitivity to the quality of Top-$k$ candidates compared to Gemma-2.

Meanwhile, under extreme noise, as seen in Table \ref{table:WRLCresult}, a Beam Width of 5 consistently led to much poorer WER for both models. This is likely due to target glosses being pushed out of the limited Top-5 search space. Conversely, higher Beam Widths (10-15) generally improved performance, effectively balancing candidate exploration with noise introduction.  Ultimately, these evaluations consistently show that the LLM algorithm's performance is directly constrained by the quality of its input probability distributions; thus, accurate initial gloss candidate generation is salient.

\begin{table}[t]
\caption{Comparison at Different Word Replacement and Distribution Corruption Rates}
\centering
\label{table:WRresults}
\begin{tabular}{@{}cccccc@{}}
\toprule
 & & \multicolumn{2}{c}{Phi-3 Mini $_{\text{\textit{(WER)}}}$} & \multicolumn{2}{c}{Gemma-2 9B $_{\text{\textit{(WER)}}}$} \\ \cmidrule(l){3-6}
\multirow{-2}{*}{WR} & \multirow{-2}{*}{DC} & Top-1 $\downarrow$     & Top-5 $\downarrow$& Top-1 $\downarrow$      & Top-5 $\downarrow$\\ \midrule
\rowcolor[HTML]{E5E4E2}
\multicolumn{6}{c}{\textit{Fixed Distribution Corruption}} \\
25\%  & 5  & 0.2446   & 0.1981 & 0.1799   & 0.1316 \\
50\%  & 5  & 0.3035   & 0.2385 & 0.2555   & 0.1714 \\
75\%  & 5  & 0.4667   & 0.3783 & 0.3420   & 0.2648 \\
100\% & 5  & 0.6231   & 0.5036 & 0.4807   & 0.3416 \\ 
\midrule
\rowcolor[HTML]{E5E4E2}
\multicolumn{6}{c}{\textit{Fixed Word Replacement}} 
\\
50\%  & 5  & 0.3192   & 0.2470 & 0.2661   & 0.1880 \\
50\%  & 10  & 0.3648   & 0.2562 & 0.2687   & 0.2156 \\
50\%  & 15  & 0.4065   & 0.3453 & 0.3205   & 0.2517 \\
50\% & 20  & 0.4396   & 0.3867 & 0.4086   & 0.3495 \\ \bottomrule
\end{tabular}
\end{table}

\begin{table}[t]
\caption{Table of Results under Extreme Noise Scenarios}
\centering
\label{table:WRLCresult}
\begin{tabular}{@{}ccccccc@{}}
\toprule
 & & & \multicolumn{2}{c}{Phi-3 Mini $_{\text{\textit{(WER)}}}$} & \multicolumn{2}{c}{Gemma-2 9B $_{\text{\textit{(WER)}}}$} \\ \cmidrule(l){4-7} 
\multirow{-2}{*}{WR} & \multirow{-2}{*}{DC} & \multirow{-2}{*}{BW} & Top-1 $\downarrow$      & Top-5 $\downarrow$   & Top-1 $\downarrow$      & Top-5 $\downarrow$     \\ \midrule
50\%  & 30 & 5  & 0.4898   & 0.4835 & 0.4634   & 0.4378 \\
50\%  & 30 & 10  & 0.3843   & 0.3420 & 0.3016   & 0.2334 \\
50\%  & 30 & 30  & 0.4091   & 0.3207 & 0.3776   & 0.3136 \\
100\% & 30 &  5  & 0.8313   & 0.7945 & 0.8274   & 0.8052 \\ 
100\% & 30 &  15  & 0.7376   & 0.6520 & 0.6316   & 0.5491 \\ 
100\% & 30 &  30  & 0.7820   & 0.7282 & 0.6686   & 0.6219 \\ 
100\% & 30 &  50  & 0.6080   & 0.5054 & 0.6915   & 0.5789 \\ 
\bottomrule
\end{tabular}
\end{table}

\subsection{\textbf{Full System Evaluation}} \label{sec:full sys eval}

Following validation on synthetic data, we integrate the disambiguation module into the sign spotter for real-world evaluation. A key challenge, however, is the scarcity of public datasets offering both gloss-annotated continuous sign language and a paired isolated sign video dictionary. While datasets like Phoenix14T \cite{8578910}, CSLDaily \cite{9578398}, and MeinDGS \cite{hanke-etal-2020-extending} provide continuous glosses, they lack isolated video dictionaries. Conversely, BSLDict \cite{varol2022scale} offers dictionaries without continuous gloss-annotated video.

Therefore, we leverage an internally collected continuous sign language dataset, and pair it with a dictionary of 1000 vocabulary entries, each associated with an isolated sign video. We then conduct evaluations in two parts, first evaluating all previously discussed fusion methods without disambiguation against a baseline (Sec \ref{performance without bs}), then assessing the impact of integrating the LLM disambiguation module (Sec \ref{performance bs}).

\subsubsection{\textbf{Performance of Sign Spotters (No Disambiguation)}} \label{performance without bs}
Table \ref{table:CSLRresult} presents the performance of various sign spotter configurations without the disambiguation module. We employ the I3D Sign Spotter (without dictionary-matching) \cite{sincan2024using} as a baseline, and it performed worst with WER of 0.9089. This high error rate stems from its fixed 2,281-gloss BOBSL vocabulary, leading to OOV issues that are impractical to mitigate through retraining due to data scarcity.

In contrast, approaches utilizing the dictionary-matching algorithm achieved substantially lower WERs (around 0.5). This highlights the benefits of dictionary-matching, which offers superior vocabulary flexibility and a training-free nature that allows easy incorporation of new gloss entries. Among these, the \textbf{\textit{Late Fusion}} and \textbf{\textit{Full-Ensemble}} methods proved most effective, achieving WERs of 0.4724 and 0.4924, respectively. Their improved performance is attributed to feature fusion techniques that enhanced discrimination for dictionary matching. Notably, the Late Fusion approach consistently performed best, aligning with its highest Top-1 Gloss Accuracy (54.51\%) observed during $\alpha_{\text{late}}$ tuning (Table \ref{table:tune1}).

\begin{table}[b]
\caption{Result Comparisons for Different Fusion Methods without Disambiguation Module}
\centering
\label{table:CSLRresult}
\begin{tabular}{@{}ccc@{}}
\toprule
 Approach & Top-1 $\downarrow$ $_{\text{\textit{(WER)}}}$\\ 
 \midrule
BOBSL I3D Sign Spotter \cite{sincan2024using} & 0.9089 \\
I3D (DTW+Cosine Fusion)  & 0.5117  \\
RH (DTW+Cosine Fusion)  & 0.6909  \\
\textbf{RH+I3D Late Fusion }& \textbf{0.4724}  \\
I3D+RH+LH Intermediate Fusion & 0.5060  \\
IF+I3D Full-Ensemble & 0.4924  \\ 
\bottomrule
\end{tabular}
\end{table}

\begin{table}[b]
\caption{Result Comparisons with Disambiguation Module}
\centering
\label{table:FinalBSresult}
\begin{tabular}{ccccc}
\toprule
 & \multicolumn{2}{c}{Phi-3 $_{\text{\textit{(WER)}}}$} & \multicolumn{2}{c}{Gemma-2 $_{\text{\textit{(WER)}}}$} \\ \cmidrule(l){2-5} 
\multirow{-2}{*}{Approach} & Top-1 $\downarrow$  & Top-5 $\downarrow$ &Top-1 $\downarrow$    & Top-5 $\downarrow$\\ \midrule
\textbf{RH+I3D Late Fusion}  & \textbf{0.4438}   & \textbf{0.3562} & \textbf{0.4473}   & \textbf{0.3481} \\
IF+I3D Full-Ensemble & 0.4567   & 0.3779 & 0.4647   & 0.3644 \\
\bottomrule
\end{tabular}
\end{table}

\subsubsection{\textbf{Impact of LLM Beam Search Integration}} \label{performance bs}

Integrating the disambiguation module into the top two performing dictionary-matching approaches led to further performance improvements, as seen in Table \ref{table:FinalBSresult}. Specifically, the Late Fusion WER decreased from 0.4724 to 0.4438 (Phi-3) and 0.4473 (Gemma-2); while the Ensemble method's WER dropped from 0.4924 to 0.4567 (Phi-3) and 0.4647 (Gemma-2). This highlights the disambiguation module's ability to incorporate linguistic knowledge when forming the final sequence of glosses, as it moves beyond a simple output of individually spotted glosses. Additionally, utilizing beam search also yields multiple highly probable gloss combinations, increasing the likelihood of obtaining accurate gloss sequences. This benefit persists even if the top-ranked combination is imperfect, as the correct sequence may still reside among the lower-ranked alternatives.

\subsection{Qualitative Results}

This section presents qualitative results visualizing the performance of our best-performing approach: the \textbf{Late Fusion with integrated disambiguation module}. We first illustrate its Top-3 gloss predictions against the ground truth (Sec \ref{top3performance}). Then, we provide comparisons between the baseline I3D sign spotter and the Late Fusion model, both with and without the disambiguation module, to highlight the impact of our proposed components (Sec \ref{finalspotperformance}).

\subsubsection{\textbf{Top-3 vs Ground-Truth}} \label{top3performance}

\begin{table}[b]
\footnotesize
\caption{Comparison of Top-3 predictions vs Ground Truth}
\label{topthree}
\begin{tabular}{l|l}
\hline \hline
\rowcolor[HTML]{dae8fc}
\textbf{Ground Truth:}                      & I LOVE WALKING SUNDAY AFTERNOON                                                                                 \\ 
\textbf{Top-3 Prediction:} & 1. ME LOVE WALKING SUNDAY AFTERNOON                                                                             \\
& \textbf{2. I LOVE WALKING SUNDAY AFTERNOON}                                                                     \\
& 3. MYSELF LOVE WALKING SUNDAY AFTERNOON                                                                         \\ \hline \hline
\rowcolor[HTML]{dae8fc}
\textbf{Ground Truth:}                      & NAME ME FS\_CHRIS FS\_WOOD                                                                                      \\ 
\textbf{Top-3 Prediction:} & 1. NAME MY FS\_CHRIS FS\_WOOD \\
& 2. NAME MY FS\_CHRIS JUNE                                                                                       \\
& \textbf{3. NAME ME FS\_CHRIS FS\_WOOD}                                                                          \\ 
\hline \hline
\rowcolor[HTML]{dae8fc}
\textbf{Ground Truth:}                      & \begin{tabular}[c]{@{}l@{}}NEXT TRAIN ARRIVE PLATFORM PEOPLE\\ BOARD IMPOSSIBLE\end{tabular}                    \\ 
\textbf{Top-3 Prediction:} & \begin{tabular}[c]{@{}l@{}}1. NEXT RAIL ARRIVE PLATFORM CUSTOMER\\ BOARD IMPOSSIBLE\end{tabular}                \\
& \textbf{\begin{tabular}[c]{@{}l@{}}2. NEXT TRAIN ARRIVE PLATFORM CUSTOMER\\ BOARDING IMPOSSIBLE\end{tabular}}   \\
& 3. NEXT RAIL ARRIVE PLATFORM ACKNOWLEDGE \\ & BOARDING IMPOSSIBLE \\
\hline \hline
\end{tabular}
\end{table}

An analysis of Top-3 predictions against ground truth in Table \ref{topthree} demonstrates the disambiguation approach's accuracy. Predictions were generally strong, but the most precise sequences often appeared at ranks 2 or 3, particularly when glosses were semantically similar. For instance, in Example 1, the Top-3 outputs were near-identical, with differences only in the glosses ``ME'', ``I'', and ``MYSELF''. This confusion stems from the significant semantic and signing motion overlap between these signs. Similar ambiguities were also observed with ``ME''/``MY'' (Example 2) and ``TRAIN''/``RAIL'' (Example 3). Fortunately, such instances typically result in minimal translational discrepancies, thus a contextually sound option is often present within the Top-3 candidates.

\subsubsection{\textbf{Qualitative Comparisons against Baseline}} \label{finalspotperformance}
Additionally, we also offer a qualitative comparison between the baseline I3D Sign Spotter and our proposed method in Table \ref{tab:compare w baseline}. Here, the baseline spotter consistently produced highly inaccurate gloss sequences. This deficiency likely stems from its fixed vocabulary, which can lead to severe OOV issues and produce outputs which are frequently irrelevant to the ground truth. In contrast, our dictionary-based approach demonstrated substantially improved performance, capitalizing on the the vocabulary flexibility afforded by the training-free dictionary and the enhanced gloss classification via feature fusion. The subsequent integration of the disambiguation module (Ours w/ LLM) then further refined these outputs. For instance, it corrected ``US'' to ``WE-ALL'' (Example 2, 4) and ``SHORT" to ``MAKE" (Example 3), clearly demonstrating its error-correcting capabilities.

Despite these improvements, failure cases were present, primarily stemming from semantic ambiguities between predicted and target glosses (e.g., ``DIVERSITY'' or ``VARIETY'' for ``MULTIPLE-DIFFERENT'' in Example 2). Although such inaccuracies elevate WER, the resulting translations generally still remain comprehensible. Thus, the LLM integration provides significant contributions, its value being evident even with these identified limitations.

\begin{table}[t]
\footnotesize
\caption{Qualitative Comparisons against Baseline Method}
\label{tab:compare w baseline}
\begin{tabular}{l|l}
\hline \hline
Ground Truth                       & I WORK UNIVERSITY                                                                                                       \\
Baseline \cite{sincan2024using}            & HEART WALL UNIVERSITY                                                                                          \\
\textbf{Ours}                   & \textbf{I WORK UNIVERSITY}                                                                                                       \\
\textbf{Ours w/ LLM} & \textbf{I WORK UNIVERSITY}                                                                                                       \\ \hline \hline
Ground Truth                       & \begin{tabular}[c]{@{}l@{}}WE-ALL PROVIDE SKILLS GENIUS\\ MULTIPLE-DIFFERENT AREA\end{tabular}                                                                 \\
Baseline \cite{sincan2024using}              & NEED SECOND WARM MOISTURE EXPLAIN EXPLAIN                                                                               \\
\textbf{Ours}                   & US \textbf{PROVIDE SKILLS GENIUS} VARIETY \textbf{AREA}                                                                                 \\
\textbf{Ours w/ LLM} & \textbf{WE-ALL PROVIDE SKILLS GENIUS} DIVERSITY \textbf{AREA}                                                                             \\ \hline \hline
Ground Truth                       & TOGETHER WE MAKE CHANGE BEEN                                                                                            \\
Baseline \cite{sincan2024using}              & EXPECT MAKE GO MANY MANY                                                                                                \\
\textbf{Ours}                   & \textbf{TOGETHER WE} SHORT \textbf{CHANGE} PAIN                                                                                         \\
\textbf{Ours w/ LLM} & \textbf{TOGETHER WE MAKE CHANGE} PAIN                                                                                            \\ \hline \hline
\textbf{Ground Truth}                     & \begin{tabular}[c]{@{}l@{}}WE-ALL IDEAS BOTH-FORWARD OPEN-MINDED\\ IN EVERYTHING WE-DO\end{tabular}                     \\
Baseline \cite{sincan2024using}             & \begin{tabular}[c]{@{}l@{}}SECOND WALK MOISTURE EXPLAIN\\ REASON WELL GO \end{tabular} \\
\textbf{Ours}                   & \begin{tabular}[c]{@{}l@{}}US \textbf{IDEAS BOTH-FORWARD OPEN-MINDED}\\ AFFECT \textbf{EVERYTHING WE-DO}\end{tabular}                     \\
\textbf{Ours w/ LLM} & \begin{tabular}[c]{@{}l@{}}\textbf{WE-ALL IDEAS BOTH-FORWARD OPEN-MINDED}\\ AFFECT \textbf{EVERYTHING WE-DO}\end{tabular}                 \\ \hline \hline
\end{tabular}
\end{table}

\section{Conclusion}
In this work, we introduced a novel framework for sign spotting that integrates an LLM and uses a modified beam search decoding approach. By combining LLM-derived linguistic priors with visually grounded emission probabilities, our approach effectively disambiguates gloss sequences. The efficacy of this method was demonstrated by a significant reduction in Top-1 WER on synthetic data from 100\% to under 63\%. Meanwhile, on real-world sign language videos, our dictionary-matching system, coupled with I3D and ResNeXt101 feature fusion, lowered Top-1 WER from 90.89\% to 47.25\%. The subsequent integration of the LLM disambiguation module led to further improvements, achieving a Top-1 WER of 44.73\% and a Top-5 WER of 34.81\%. Qualitative analyses complemented these quantitative gains, highlighting clear advantages over baseline methods. While acknowledging limitations such as dictionary noise, our findings strongly suggest that sign spotting approaches can be advanced through the introduction of linguistic contexts.

\vspace{1em}

\noindent\textbf{Acknowledgments:} 
This work was supported by the SNSF project ‘SMILE II’ (CRSII5 193686), the Innosuisse IICT Flagship (PFFS-21-47), EPSRC grant APP24554 (SignGPT-EP/Z535370/1) and through funding from Google.org via the AI for Global Goals scheme. This work reflects only the author’s views and the funders are not responsible for any use that may be made of the information it contains.

\bibliographystyle{ACM-Reference-Format}
\bibliography{main}

\end{document}